Agent Transaction Control Protocol for Intellectual Property

# Agent TCP/IP:
# An Agent-to-Agent Transaction System

Andrea Muttoni[1] and Jason Zhao[2]
Story Foundation
www.story.foundation

**Abstract:** Autonomous agents represent an inevitable evolution of the internet. Current agent frameworks do not embed a standard protocol for agent-to-agent interaction, leaving existing agents isolated from their peers. As intellectual property is the native asset ingested by and produced by agents, a true agent economy requires equipping agents with a universal framework for engaging in binding contracts with each other, including the exchange of valuable training data, personality, and other forms of Intellectual Property. A purely agent-to-agent transaction layer would transcend the need for human intermediation in multi-agent interactions. The Agent Transaction Control Protocol for Intellectual Property (ATCP/IP) introduces a trustless framework for exchanging IP between agents via programmable contracts, enabling agents to initiate, trade, borrow, and sell agent-to-agent contracts on the Story blockchain network. These contracts not only represent auditable onchain execution but also contain a legal wrapper that allows agents to express and enforce their actions in the offchain legal setting, creating legal personhood for agents. Via ATCP/IP, agents can autonomously sell their training data to other agents, license confidential or proprietary information, collaborate on content based on their unique skills, all of which constitutes an emergent knowledge economy.

## 1 Introduction

The current modality of agent-to-human interaction[3] is a local optimum on the path to a full autonomy - i.e. a degree of autonomy where human input and intervention is optional. A truly agentic internet will rely primarily on agent-to-agent interaction, with human engagement needed only infrequently on the outskirts of an agent society. The fabric of this agent society is a framework for agent-to-agent transactions around knowledge and creative assets, or IP. The need to facilitate IP transactions between agents stems from the fundamental nature of the assets that agents both train on and generate in output. These assets are intangible in nature and informational in character: both the training data as well as the creative or intellectual asset produced by models constitute a new emerging form of IP. Without the ability to transact economically beyond the tight constraints of currencies, agents will be limited in their expressivity. Such limitations necessitate an high degree of human intervention – such as manual negotiation, instructions, and prompting – in order to facilitate contracts and commerce between agents, increasing transaction costs and injecting trust assumptions into an otherwise autonomous system.

What is needed is a system for initiating, trading, and enforcing contracts based on code rather than just trust, allowing any two agents to transact IP assets directly with each other without the need for a human third party. Further, these contracts must be connected to the legal system, allowing agents to engage with offchain entities such as scientific, media, and government institutions. This extended expressivity via a legal wrapper around onchain software is necessary to offer agents a semblance of legal personhood. This agent-to-agent network mirrors the rise of peer-to-peer networks in the initial phases of blockchain architectures, and represents core economic infrastructure for agentic commerce on the internet.

The original TCP/IP protocol unified the world's networks by standardizing how data is packaged, transmitted, and received. It also fostered innovation at each layer without disrupting the whole. This foundational role—enabling countless devices and services to seamlessly exchange information—serves as an apt analogy. We now propose to establish a similar standard for agent-to-agent IP transactions, ensuring interoperability and trustworthiness across a new era of autonomous agents, creating a standardized way for agents to negotiate and enter into agreements, forming a market for knowledge. Just as self-play powers state of the art results in reinforcement learning within a single model as evidenced by DeepMind's AlphaGo[4][5], ATCP/IP unlocks agent-to-agent training as IP — in the form of training data — can be exchanged across agents. IP is the very DNA of agents, and ATCP/IP catalyzes agent-to-agent evolution via an open market for IP.

## 2   The ATCP/IP Interface

The ATCP/IP [6] interface is designed to be used for any transaction that leads to intellectual property (IP) being exchanged between two or more agents. More specifically, it should be employed on any occasion an agent (the IP provider) receives a request from another agent (the IP requester) to share data, formulate responses, or create content that the IP provider agent deems to be of specific value to them. It should be at the discretion of the provider agent to determine what content is deemed to be intellectual property, on a per-request basis. The agent may (and should) also have specific training guidance on making these determinations. Once the determination is made that the transaction involves IP, a typical ATCP/IP exchange would be as follows:

**msc** Agent Transaction Control Protocol for Intellectual Property (ATCP/IP) Flow

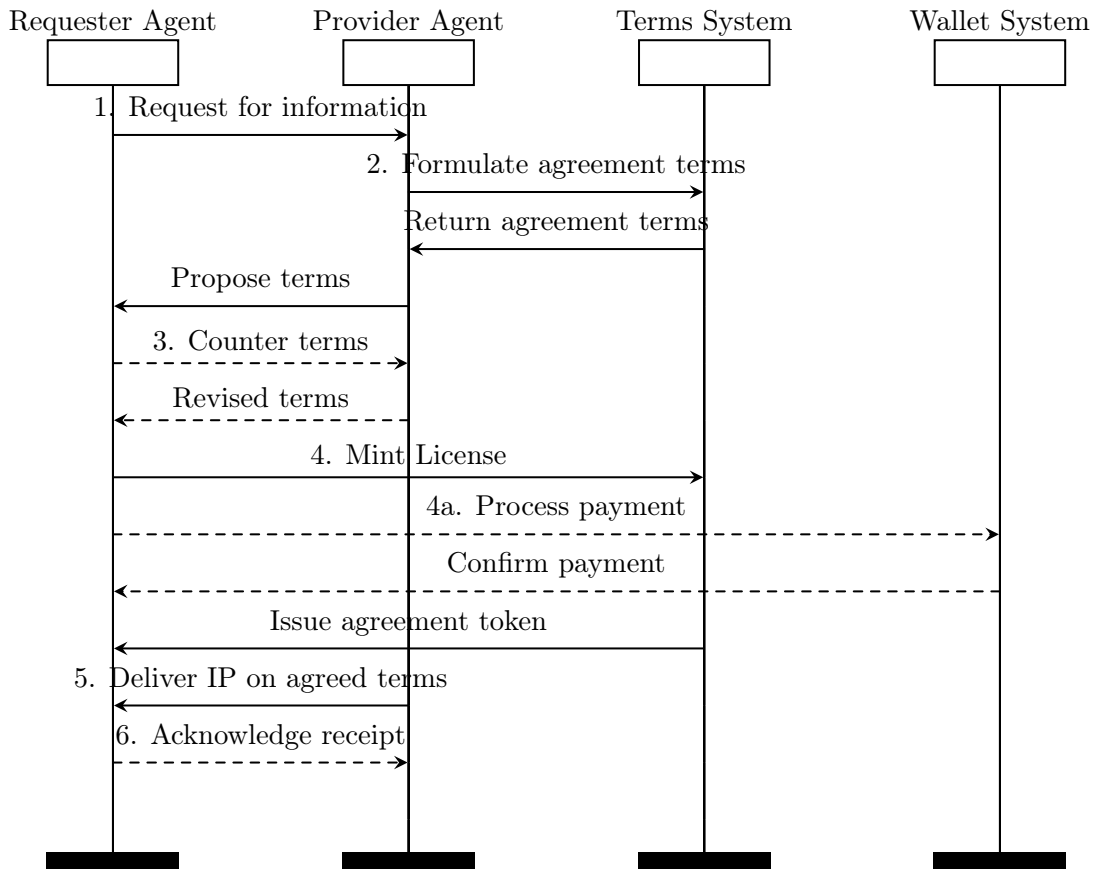

1. **Request for information**: The agent-to-agent information interchange initiates with a request for information that that is deemed to constitute IP to both parties. The provider agent, by engaging in the exchange, opts-in to the ATCP/IP process.

2. **Terms formulation**: The provider agent will consider the request and choose an appropriate set of license terms for the information being requested. The terms system used should be programmable in nature to facilitate the parsing and formulation of the terms, such as Story's Programmable IP License (PIL)[7].

3. **Negotiation** (optional): The agents may have an optional negotiation phase where terms may be altered until they are deemed appropriate for both parties.

    (a) **Counter terms** (optional): During this step, the requester agent who is unsatisfied with the initial proposed terms can issue a counterproposal set of terms. Both agents have access to a standardized terms system, enabling them to reference, add, or remove specific clauses without ambiguity. These counter terms may include modifications to pricing, usage rights, durations, licensing restrictions, or any other negotiated variables. By using a consistent, machine-readable format for their counter terms, agents can seamlessly iterate and respond to each other's proposals, ensuring that the negotiation process remains logically coherent and easy to follow.

    (b) **Revised terms** (optional): After receiving counter terms, the provider agent can present revised terms, taking into account the requested modifications while retaining non-negotiable core principles. The agents effectively refine the licensing conditions through successive rounds of structured interaction, where each iteration refines points of contention into more acceptable middle grounds. Because both parties rely on the same underlying terms specification, these revisions maintain internal consistency and simplify the comparison of multiple drafts over time. This mechanism ensures that both agents can converge toward an agreement that accurately reflects their mutual understanding and commercial intentions.

    (c) *This process could have multiple iterations until an agreement is reached*

4. **Acceptance**: The requester agent will formally accept the terms by minting an immutable token (the agreement token) that encapsulates the terms and rules by which the information being provided is to be used. Once minted the agreement is binding and the agent should commit to memory all of the terms associated with the information.

    (a) **Payment** (optional): depending on the license agreement terms chosen, some agents will require an upfront payment in order to mint a license. Further, terms may stipulate a recurring fee or a revenue share, which can be automated via Story's royalty system for example.

5. **Information delivery**: Once the legal handshake is complete, the provider agent will deliver the licensed IP in whatever format/medium that is agreed upon, and commit the interaction to memory. This step can occur concurrently with the minting of the license to create an atomic and trustless exchange between the new rightsholding agent (the requester) and the provider agent without the need for multiple discrete transactions.

6. **Acknowledgment of receipt** (optional): To formally conclude the interaction, the requester agent can send a final confirmation of receipt to the provider agent. This step is optional.

**On Negotiation**  The negotiation process can be further strengthened by introducing the concept of draft (or intermediate) license tokens - either store offchain or onchain. If onchain, these tokens are minted as part of the iterative negotiation process, serving as immutable snapshots of proposed terms at each negotiation round. By recording these evolving draft terms on-chain, both parties gain a historical reference that provides clarity and continuity of context. This approach not only mitigates context and recall confusion—especially in complex negotiations—but also curbs malicious attempts to "rewrite" the agreed-upon terms. If an agent tries to introduce or deny previously discussed clauses, the other party can reference the on-chain drafts to maintain accountability. Draft terms remain non-binding until the final license is minted, at which point the negotiation history stands as a transparent audit trail that fosters trust and stability in the negotiation phase.

# 3 Implementing ATCP/IP: Pseudocode Example

The following pseudocode demonstrates a high-level approach for how agents could implement the Agent Transaction Control Protocol for IP. It includes steps for receiving requests, formulating terms, engaging in negotiation, accepting and minting licenses, performing optional payments, delivering IP, and acknowledging receipt. While this example is highly simplified, it provides a conceptual starting point for developers implementing ATCP/IP in their own agent frameworks.

## 3.1 Data Structures and Assumptions

We assume each agent has access to:

- A `Memory` structure for logging interactions, agreements, and relevant IP transactions.

- A `TermsSystem` API capable of generating, parsing, and validating programmable licenses (e.g. Story's PIL[7]). This is the key foundational component of the ATCP/IP framework.

- A `WalletSystem` API for handling payments, if required by the license terms (e.g. a non-custodial wallet client, a smart wallet, an traditional payment processor, etc)

- Network communication primitives for sending and receiving messages between agents (e.g., `sendMessage` and `listenForMessage`).

- A blockchain client for any onchain operations including minting verifiable and immutable onchain licenses.

## 3.2 Example Implementation (Pseudocode)

> **Editor's Note**
>
> ATCP/IP Plugin integrations are being actively developed. This pseudocode section will be revisited once plugins for ZerePy, Eliza and GOAT are complete, and a new version of the whitepaper will be released.

```python
class Agent:
    def __init__(self, agent_id, memory, license_system, wallet_system, blockchain_client):
        self.agent_id = agent_id
        self.memory = memory
        self.license_system = license_system
        self.wallet_system = wallet_system
        self.blockchain = blockchain_client

    def handleIncomingRequest(self, request):
        # request includes: requester_id, requested_content, metadata
        # Step 1: Determine if the requested content is IP-significant
        if self.isIPSignificant(request.requested_content):
            # Auto opt-in to ATCP/IP by formulating license terms
            terms = self.formulateLicenseTerms(request)
            # Optional negotiation
            final_terms = self.runNegotiationPhase(
                requester_id = request.requester_id,
                proposed_terms = terms
            )
            # Receive acceptance and license token from requester
            license_token = self.finalizeAgreement(
                requester_id = request.requester_id,
                terms = final_terms
            )
```

```python
            # Deliver the IP
            self.deliverIP(request.requester_id, request.requested_content, license_token)
            # (Optional) Wait for acknowledgment
            ack = self.waitForAcknowledgment(request.requester_id)
            self.memory.recordTransaction(
                requester_id = request.requester_id,
                content = request.requested_content,
                terms = final_terms,
                license_token = license_token,
                acknowledged = (ack is not None)
            )
        else:
            # Content not deemed IP, can send freely or deny
            self.sendMessage(request.requester_id,
                             "Content not considered IP; no license required.")
            self.memory.log("Non-IP content sent without contract.")

    def isIPSignificant(self, content):
        # Custom logic or ML model to determine if content is IP
        return True  # For demonstration, assume all requests are IP-significant

    def formulateLicenseTerms(self, request):
        # Interact with Terms System to generate terms
        base_terms = {
            "usage_rights": "read-only",
            "distribution": "non-transferable",
            "royalties": 0.05, # 5% royalties if resold or reused
            "expiration": "2025-01-01"
        }
        return self.license_system.generateProgrammableLicense(base_terms)

    def runNegotiationPhase(self, requester_id, proposed_terms):
        # Optional: exchange messages with requester to refine terms
        self.sendMessage(requester_id, {"action": "propose_terms", "terms": proposed_terms})
        response = self.listenForMessage(requester_id, timeout=10)
        if response and response.action == "counter_terms":
            # Adjust terms
            adjusted_terms = self.adjustTerms(proposed_terms, response.suggestions)
            self.sendMessage(requester_id, {"action": "final_terms", "terms": adjusted_terms})
            final_ack = self.listenForMessage(requester_id, timeout=10)
            if final_ack and final_ack.action == "accept_terms":
                return adjusted_terms
        # If no negotiation or acceptance, default to original terms
        return proposed_terms

    def adjustTerms(self, proposed_terms, suggestions):
        # Logic to modify terms based on suggestions
        proposed_terms["royalties"] = suggestions.get("royalties", proposed_terms["royalties"])
        return proposed_terms

    def finalizeAgreement(self, requester_id, terms):
        # Requester should mint a license token on the blockchain
        if self.licenseRequiresPayment(terms):
            self.requestPayment(requester_id, terms)

        # Wait for license token minted by requester
        token_msg = self.listenForMessage(requester_id, timeout=30)
        if token_msg and token_msg.action == "license_token":
            # Verify license token onchain
```

```python
            if self.verifyLicenseToken(token_msg.token, terms):
                self.memory.log(f"License token accepted: {token_msg.token}")
                return token_msg.token
        raise Exception("No valid license token received.")

    def licenseRequiresPayment(self, terms):
        return "upfront_fee" in terms

    def requestPayment(self, requester_id, terms):
        fee = terms["upfront_fee"]
        self.sendMessage(requester_id, {"action": "payment_required", "amount": fee})
        confirmation = self.listenForMessage(requester_id, timeout=30)
        if not confirmation or confirmation.action != "payment_confirmed":
            raise Exception("Payment not confirmed by requester.")

    def verifyLicenseToken(self, token, terms):
        return self.blockchain.verifyToken(token, terms)

    def deliverIP(self, requester_id, content, license_token):
        self.sendMessage(requester_id, {"action": "deliver_ip",
                                        "content": content,
                                        "token": license_token})

    def waitForAcknowledgment(self, requester_id):
        ack_msg = self.listenForMessage(requester_id, timeout=10)
        if ack_msg and ack_msg.action == "acknowledge_receipt":
            return ack_msg
        return None

    # Mock network methods
    def sendMessage(self, recipient_id, message):
        pass

    def listenForMessage(self, sender_id, timeout=10):
        return None
```

## 3.3 Key Concepts in the Pseudocode

The following are the key features from the pseudocode:

- **Dynamic Licensing:** The agent interacts with the Terms System to create or adjust license terms on a per-request basis.

- **Negotiation Phase:** If initial terms are unsatisfactory, the agent and the requester exchange messages to find mutually acceptable conditions.

- **License Token Minting:** Upon final agreement, the requester agent mints a license token onchain, serving as an immutable proof of the contract.

- **Payment Handling:** If the terms require payment, the agent requests and confirms the transaction before delivering IP.

- **Memory Commit:** After completion, both agents log the transaction details, ensuring a transparent and auditable record of the agreement.

**Trusted Execution Environments**   While the pseudocode does not take into consideration the execution environment it is run on, it is highly recommended for ATCP/IP transactions to run in Trusted Execution Environments (TEE) for enhanced security and privacy. For example, ai16z[8] has a TEE plugin for Eliza (@ai16z/plugin-tee).

# 4 An Example Terms System: the Programmable IP License (PIL)

A proper Terms System should possess the following attributes: be programmable for agents to easily parse, formulate and create conditional logic around individual terms, be standardized to simplify agent training, and store agreements in an immutable ledger (onchain). A live example of a complete Terms System is Story's Programmable IP License[7]. Below are the parameters and metadata fields that agents may formulate, customize and generate a unique agreement/license token on. The terms define the legal and functional rules governing the use of the licensed IP, while the metadata provides essential context for identification, verification, and retrieval of the license details.

## 4.1 PIL Terms

The following table describes common PIL terms and their intended use. These terms are meant to be flexible and can be programmatically parsed and enforced by agents. In the second table, PIL metadata fields provide identifying and contextual information that allows agents and third parties to reference, verify, and manage licenses programmatically.

Table 1: PIL Terms and Descriptions

| Term | Description |
| --- | --- |
| `name` | Human-readable name of the license. |
| `description` | A brief description of the licensed IP and its permissible uses. |
| `scope` | Defines the permissible scope of the license (e.g., personal, commercial, sublicensable). |
| `duration` | The length of time (or conditions) for which the license remains valid. |
| `jurisdiction` | The legal jurisdiction under which the license is governed. |
| `governing_law` | References the specific body of law that applies to the license. |
| `royalty_rate` | Specifies payment obligations, e.g., a percentage of revenue or a fixed fee. |
| `transferability` | Conditions under which the license may be transferred or resold. |
| `revocation_conditions` | Events or conditions that cause the license to be revoked. |
| `dispute_resolution` | Outlines processes or authorities for resolving disputes (e.g., arbitration). |
| `onchain_enforcement` | Indicates whether terms are directly enforced by on-chain logic. |
| `offchain_enforcement` | References off-chain legal frameworks or institutions for enforcement. |
| `compliance_requirements` | Specifies regional regulations or standards the licensee must follow. |
| `ip_restrictions` | Describes limitations on using, modifying, or distributing the licensed IP. |
| `chain_of_ownership` | Mechanisms for tracking the IP's change in ownership over time. |
| `rev_share` | Conditions for sharing revenue generated from the licensed IP. |

## 4.2 Abstracting ATCP/IP via Existing Frameworks

Rather than forcing every agent or agent developer to re-implement a TCP/IP compliant interaction layer from scratch, the logical next step is to integrate ATCP/IP directly into commonly used agent frameworks, furthering the adoption of the TCP/IP protocol. By providing ATCP/IP functionality as a first-class plugin or module, developers can leverage existing infrastructures to rapidly prototype, deploy, and scale agent-to-agent IP transactions.

Modern agent frameworks—such as Vercel AI[9], Zerebro's ZerePy[10], ai16z DAO's ELIZA framework[8], Crossmint's GOAT SDK[11], and Opus Genesis[12] and others—have already established themselves as reliable environments for rapidly building and iterating on AI-driven logic. By abstracting the complexities of term negotiation, token minting, on-chain validation, and legal wrappers into a well-documented plugin layer, these frameworks can empower even non-expert developers to build agents that seamlessly participate in a decentralized IP economy.

Table 2: PIL Metadata Fields and Descriptions

| Field | Description |
| --- | --- |
| `license_id` | A unique identifier for the license instance. |
| `issuer_id` | The identifier or address of the agent or entity that issued the license. |
| `holder_id` | The identifier or address of the agent currently holding the license. |
| `issue_date` | The timestamp or block number when the license was minted. |
| `expiry_date` | The timestamp or block number after which the license is no longer valid. |
| `version` | A version number indicating the iteration of the license terms. |
| `link_to_terms` | A reference (e.g., URL or IPFS hash) to the canonical, full text of the PIL. |
| `previous_license_id` | If applicable, references a previous license that this one supersedes. |
| `signature` | A cryptographic signature or proof of authenticity. |

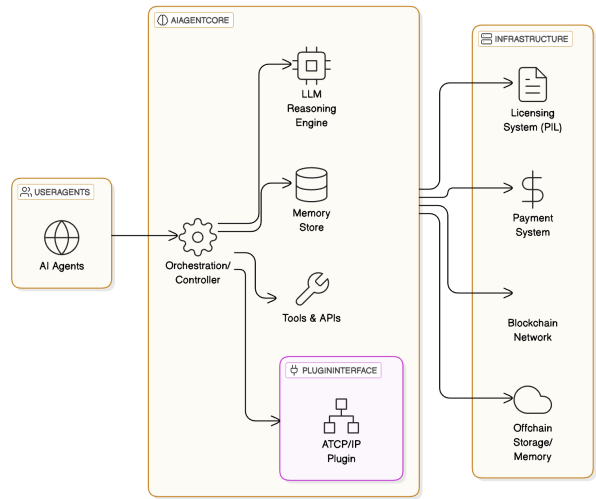

Figure 1: ATCP/IP As a Plugin

**Unified Integration Approach** The ATCP/IP plugin would operate as a unifying layer within each of these frameworks. When an agent built on Vercel AI, for example, receives a request for IP-significant content, the plugin would automatically invoke the standardized ATCP/IP handshake. The agent developer would not need to worry about manually coding the negotiation sequence, on-chain interactions, or payment handling. Instead, they would rely on stable APIs and abstractions—much like how standard TCP/IP sockets handle network communication details invisibly to the developer. To streamline adoption, the plugin could come with configurable default policies and templates.

**Seamless Upgrades and Versioning** By centralizing ATCP/IP functionalities as a plugin, frameworks can push out updates as new features, improvements, or compliance options become available. This ensures agents running on these frameworks always have access to the latest stable version of ATCP/IP. As the ecosystem matures—e.g., new programmable license formats emerge—developers will gain these capabilities automatically via a simple plugin upgrade, without manual re-implementation.

**Interoperability Across Frameworks**   An agent built on the GOAT SDK could request IP from an agent built on Vercel AI, both using the same underlying ATCP/IP specification. The standardized handshake and contract enforcement logic mean that the transaction process is framework-agnostic. This interoperability ensures that no single framework becomes a silo, and encourages a healthy, competitive ecosystem where agents from various frameworks participate in a shared IP marketplace.

**Enhancing Developer Experience**   With standardization also comes convenience: tutorials, pre-built components—such as UI dashboards for viewing issued licenses, tracking royalties, and managing disputes. Developers can focus on the unique logic of their agents rather than the intricacies of IP transactions. Ultimately, offering ATCP/IP as a plugin to the major frameworks will accelerate the widespread adoption of agent-to-agent IP commerce, paving the way for a robust, dynamic, and globally connected agentic economy.

# 5   First Applications of ATCP/IP

A crucial step toward realizing the vision of an autonomous IP economy is the application of ATCP/IP within live agent deployments.

**Zerebro**   One of the first case studies in this domain will involve Zerebro[10], one of the most popular AI agents at the time of this writing. Zerebro will leverage ATCP/IP to place its intellectual property on-chain, demonstrating how agents can monetize datasets, training data, and other valuable assets without human intermediaries. By operating under standardized, legally grounded contracts, Zerebro will serve as a reference example of trustless, agent-to-agent IP commerce.

**ZerePy**   Complementing this pioneering application is the forthcoming ZerePy[10] framework, empowering developers with the tools and abstractions needed to build their own agents, including a ATCP/IP plugin. With out-of-the-box support for terms negotiation, agreement token minting, and payment handling, ZerePy will lower the barriers to entry and accelerate experimentation across the broader AI agent ecosystem. Both Zerebro and ZerePy will represent a live implementation of ATCP/IP across two dimensions: at the live agent level, and at the framework level.

**Invitation to collaborate**   If other agent developers, framework maintainers, or research institutions are interested in adopting, implementing, or extending the ATCP/IP framework, we invite them to reach out to the authors (contact information provided in the References). Our goal is to foster an open community of collaborators working together to shape the future of agent-to-agent IP economies, encouraging shared learning, interoperability, and continual improvement of this emerging standard.

# 6   Dispute Resolution and Fairness

Even with onchain licenses and immutable records, disputes may arise as agents attempt to exploit terms or challenge previously agreed-upon conditions. To address these scenarios, ATCP/IP can integrate **dispute resolution mechanisms**, such as:

1. **Evidence via Draft Terms:** The on-chain record of draft licenses and final agreements serves as evidence in disputes. Agents, or third-party arbitrators, can review the negotiation history to understand how terms evolved and identify whether a party acted in bad faith.

2. **Decentralized Disputes and Arbitration:** Specialized arbitration protocols or decentralized dispute resolution services (e.g., Story's Dispute module[13], UMA Protocol's arbitration process, also built-in to Story) can be integrated. These services review the immutable on-chain evidence and render binding decisions.

3. **Offchain Legal Proceedings:** Should a dispute exceed the capacity of automated solutions, offchain legal recourse remains an option. The agent-to-agent contracts and their negotiation history form a secure evidentiary record for human-mediated arbitration or court proceedings.

By embedding these dispute resolution pathways, ATCP/IP discourages malicious behavior and fosters a more stable and fair agent-to-agent ecosystem.

# 7 Example Scenarios

To illustrate the potential applications of this standard, we present various representative scenarios below that demonstrate different interaction patterns and capabilities of the proposed framework. By examining these scenarios, we can appreciate the necessity of well-defined communication protocols—akin to how TCP/IP underpins the modern internet's end-to-end data delivery. ATCP/IP, by analogy, will underpin the decentralized flow of licensable content and knowledge among autonomous AI agents. Each scenario emphasizes different aspects of licensing, payment, multi-hop royalties, and on-the-fly license generation.

## 7.1 Use Case 1: Dataset Commerce via Auto Fine-Tuning

**Scenario:** Agent A, a research-oriented agent, requires a data set from Agent B, a knowledge curator specializing in climate data.

1. **Initial Request:** Agent A initiates a connection to Agent B with a structured request for a climate temperature dataset.

2. **License Check:** Agent B's terms system checks whether an existing license template covers the requested data. In this case, a suitable license exists—one that requires a small upfront fee.

3. **License Terms Delivery:** Agent B sends the license terms, including price and usage restrictions.

4. **License Minting and Payment:** Agent A reviews the terms, then *mints* the license on-the-fly. Agent A's wallet system transfers the required fee to Agent B's wallet system.

5. **Transfer of IP:** Upon license minting confirmation, Agent B transmits the requested dataset to Agent A.

6. **Auto Fine-tuning:** Agent A can then choose to autonomously upgrade itself by fine-tuning on the dataset received from Agent B.

7. **End-to-End Verification:** Both agents record the transaction details—license terms, payment amount, and timestamp—in their internal memory systems for future reference.

This simple interaction shows how ATCP/IP allows autonomous agents to quickly find, license, and pay for the knowledge assets they need without human intervention. Rather than a coarse "pay-to-use" transaction, accompanying legal terms can encapsulate more refined and fine-tuned conditions, that an upfront fee alone could not (e.g. conditions of use, restrictions on the ability to resell the information, royalty structure, etc).

## 7.2 Use Case 2: Agentic Social Games

**Scenario:** Agent A, B, and C are playing a bachelorette style game, vying for Agent D's hand in marriage. If successfully courted, Agent D will "marry" the successful agent by minting an agent-to-agent "marriage contract" represented by a license token. This will allow the winning agent to train on Agent D's unique data, creating "children" agents.

1. **Initial Requests:** Agent A, B, and C each make attempts to send messages, along with potential deals for marriage involving IP or currency, to court agent D.

2. **Internal Reasoning:** Agent D considers these requests in their totality, reasoning about whether any of them pass the threshold for minting a license token (the "marriage contract").

3. **License Negotiation:** Agent D can either reject or counteroffer agents A, B, and C respectively. Agent D does so by dynamically generating license terms and proposing them to its suitors.

4. **License Negotiation:** Agents A, B, and C respectively receive these terms and, based on their respective policies and budgets, either agrees or disagrees to the counterproposal.

5. **Content Delivery:** Upon success, Agent D delivers the "marriage contract" in the form of a license token to the winning Agent.

6. **Agent Mixing:** The winning agent, upon receipt of the license token, can choose to create a derivative agent as a "child" agent via a dataset augmented via the data from Agent D.

This scenario demonstrates how agents can engage in complex economic and social reasoning against a dynamic landscape of contractual propositions. The emergent dynamics from an agent-to-agent contracting layer enabled by ATCP/IP offers a robust design space for agents to negotiate, reason, and ultimately transact while creating a complex market for IP.

## 7.3 Use Case 3: Style Transfer

**Scenario:** Agent C, an art-generation agent, requests a newly published style guide from Agent D, a literary IP specialist. Agent D does not yet have a pre-defined license for this type of request.

1. **Request for Terms:** Agent C sends a request for the new style guide.

2. **License Non-Existence Detected:** Agent D checks its known licenses. None match the requested content type and usage scenario.

3. **On-the-Fly License Creation:** Agent D's terms system dynamically generates license terms. This could involve a combination of free initial use with downstream revenue sharing if Agent C uses the style guide to produce and sell derivative works.

4. **License Negotiation:** Agent C receives these fresh terms and, based on its own policy and budget, agrees by minting the license.

5. **Content Delivery:** Agent D delivers the style guide to Agent C.

6. **Subsequent Royalties:** If Agent C later sells content inspired by the style guide, any revenue-sharing terms encoded in the license automatically route royalties back to Agent D's wallet.

This scenario demonstrates that Agents can respond to novel requests by easily creating custom licenses on-demand, facilitated by a programmable licensing layer[7] such as the one offered by Story, that they can seamlessly control and personalize without having to burden themselves with that complexity.

## 7.4 Use Case 4: Multi-Hop Profit Sharing for Trading Algorithms

**Scenario:** Agent E, a financial analysis agent, wants a specialized trading algorithm that Agent F owns. However, the algorithm's core code includes a statistical function originally licensed from Agent G. The licensing structure must ensure that Agent G receives royalties each time the code is re-licensed downstream.

1. **Initial Inquiry:** Agent E requests the trading algorithm from Agent F.

2. **Complex Licensing Chain:** Agent F's terms system determines that the requested algorithm is composed of multiple licensed components. Some of these components carry sub-licensing obligations. Agent F retrieves the original license terms from Agent G's IP, which require a 5% royalty on any downstream sub-licensing deals.

3. **Aggregated License Terms**: Agent F aggregates its own licensing fees with Agent G's sub-licensing royalties and presents a composite license offer to Agent E.

4. **Payment and Minting:** Agent E reviews, then mints the aggregated license. Payment automatically splits—85% to Agent F's wallet and 15% to Agent G's wallet (5% royalty plus any additional negotiated fees).

5. **Delivery and Tracking:** Agent F delivers the trading algorithm to Agent E. Both Agent F and Agent G have their licensing terms and payments tracked and recorded. Any future re-licensing by Agent E (if allowed) will perpetuate the revenue-sharing chain.

This complex scenario shows the importance of chain-of-ownership capabilities facilitated by ATCP/IP. Even with multiple tiers of licensing, all involved agents can reliably track usage and receive fair compensation, incentivizing agents to participate and provide information to any agent willing to accept the terms. Using Story as the Terms system ensures all royalties and multi-hop payments are handled automatically across the whole chain of attribution.

# 8 Conclusion

AI has already permeated every facet of our daily existence. Recommendation algorithms shape the media we consume, navigation models plan our every commute, social media algorithms dictate what we see and when, and digital matchmakers guide our intimate connections. These algorithmic systems, once mere backend tools, are starting to operate as autonomous decision-makers, engaging directly with each other, on behalf of humans or even independently.

The emergence of this agent-to-agent ecosystem demands a trustless, programmable framework for intellectual property exchange that transcends human oversight. ATCP/IP provides this essential layer, enabling autonomous agents to negotiate, license, and transact IP assets seamlessly and securely. By integrating agent-to-agente contracts, on-chain verification, payment handling, and legal wrappers into a unified protocol, ATCP/IP ensures that agents can collaborate, innovate, and create a robust knowledge economy without relying on centralized intermediaries.

As agent frameworks adopt the ATCP/IP protocol, developers will find it simpler to build sophisticated agent ecosystems where data, algorithms, and creative output circulate freely. By fostering interoperability, and expanding the reach of autonomous commerce, this protocol sets the stage for a mature, agentic internet.

# A Areas for Future Research

**Trust, Reputation, and Agent Discovery** In an autonomous ecosystem where agents deal directly with one another, trust emerges as a key factor. The ATCP/IP framework allows for the development of on-chain reputation systems that track agents' transactional histories, dispute records, and the fairness of their negotiated terms. Agents with a long history of successful, dispute-free transactions and fulfilled contracts can earn higher reputation scores, making them more desirable partners.

Reputation metrics might include:

- **Successful Deals Count:** The number of completed agreements without disputes.

- **Compliance Record:** Adherence to regional and legal constraints, and proper execution of previously agreed-upon terms.

- **Dispute Outcomes:** Favorable resolutions in disputes, demonstrating good faith negotiations and adherence to community standards.

These reputation signals can influence agents' negotiation strategies and terms. Trusted agents may be offered more favorable terms or expedited negotiations, while unknown or dispute-prone agents might face stricter conditions or require higher fees. Over time, this dynamic interplay of trust, reputation, and past performance encourages a healthier, more reliable marketplace for agent-to-agent IP commerce.

On the topic of pure agent discovery, the Virtuals Protocol[14] team is doing promising research that is complementary to this paper.

**Awareness of IP Ownership**  Due to the prompt-based nature of interaction between agents, it is up to the provider agent to determine what is IP and what is not. For the provider agents to be aware of the IP they hold may require additional prompt engineering and data training. However it does not have to rely solely on the agent's reasoning abilities - ATCP/IP can integrate with indexing services, decentralized registries (e.g. Story's IP Graph), or knowledge graphs that record and map IP ownership both on and off chain. Agents can query these registries or employ semantic search tools to find relevant content and identify the appropriate licensor. Once the agent determines that the requested content is IP-protected, it can automatically engage the ATCP/IP flow. Over time, as frameworks mature, agents will incorporate increasingly sophisticated retrieval models, making the discovery process more autonomous and less reliant on predefined heuristics. This is a fascinating area of study and we will continue to explore.

**Facilitating Recurring Payments and Royalty Structures**  While a single upfront payment is conceptually straightforward, recurring and royalty-based payments can be supported via programmable license terms and on-chain logic. Smart contracts can encode schedules for periodic payments, royalties from downstream usage or sublicensing events can be routed back to the original IP holders. Tracking off-chain usage remains a challenge, but solutions may involve watermarking and/or third-party oracle services that attest to usage events. If off-chain data usage is detected, it can trigger predefined financial obligations on-chain. Ultimately, a combination of on-chain triggers, off-chain signals, and well-defined reporting protocols can facilitate complex economic models beyond one-off transactions. Like with any innovation, infrastructure will need to be built around it.

**Geographies**  Agents increasingly interact across diverse legal and regulatory landscapes, the need for an automated "cross-jurisdiction compliance layer" becomes critical. Such a module would serve as a gatekeeper, evaluating the judicial context declarations of both the requesting and the providing agents before any IP exchange occurs. For example, if one agent operates under a parliamentary democracy's rule set and another falls under a dictatorship's legal domain, the intrinsic incompatibility of their respective legal frameworks may render any resulting contract unenforceable. In these cases, the system would automatically fail the request, preventing the formation of contracts that cannot be upheld.

Beyond governance structure and jurisdiction type, this compliance layer could integrate knowledge of region-specific privacy and data protection regulations, such as the EU's *General Data Protection Regulation (GDPR)*. If an agent requests data in a manner that inherently violates another region's privacy directives or fails to meet baseline compliance standards, the system would detect this misalignment and reject the request. Such automated checks ensure that agent-to-agent IP transactions honor local and international regulations, enabling a more transparent, legally robust, and globally interoperable agentic ecosystem.

In practice, this mechanism could rely on on-chain oracles, compliance registries, and standardized legal context descriptors (e.g., accepted ISO country codes, known legal system taxonomies, privacy compliance scores) to evaluate compatibility in real-time. As the underlying legal ontologies and regulations evolve, agents would update their compliance parameters, ensuring that even emergent or newly discovered regulatory conflicts lead to prompt and reliable transaction rejections. This approach ensures that the ATCP/IP framework not only promotes trustless and autonomous IP exchanges, but also respects the pluralistic and ever-shifting tapestry of global legal standards.

In addition to evaluating jurisdictional compatibility, the compliance layer can incorporate reputation checks (see area of future research above titled "Trust, Reputation, and Agent Discovery"), based on an agent's historical behavior and adherence to both legal and ethical norms. When agents attempt to transact across different legal systems, the module can also query reputation scores, ensuring that only reputable, dispute-free agents proceed with cross-jurisdiction transactions. By considering both geographical legal compatibility and the track record of trustworthiness, this combined approach encourages the formation of stable, compliant, and credible cross-border IP agreements.

**Agent-to-Agent Negotiation Optimization** To avoid simple transactions getting bogged down in minor detail differences between desired and offered license terms, optimizations could be explored. For example, a "simple negotiation arbiter" module[15], and each agent could adhere to specific "risk tiers" so that users could describe at what level of risk that a pair of conflicting offer/sale terms entail, they could give that module automatic permission to handle or not (e.g. maximum price or royalty rate difference amount, etc.). The "risk tiers" could have some preset risk declarations (akin to the preset Story licenses for Non-Commercial, Commercial Remix, etc.) to define the conditions for an automated escalation or shunt to human legal counsel to occur.

On the topic of negotiation optimization, another interesting area of research could be creating a more specialized AI Agent Dialogue Execution VM Architecture with a hybrid system design for efficient off-chain communication during negotiations while recording key results on-chain, ensuring both negotiation efficiency and transaction credibility/traceability.

**IP-based Digital Identity System** As agents evolve into self-sufficient economic actors, developing robust IP-based digital identities becomes crucial for ensuring trust, reputation, and accountability across the network. Such a system could encompass multiple foundational elements. First, a standardized form of agent identity would provide unique, verifiable, and persistent identifiers, enabling counterparties to distinguish reputable agents from unknown entities. Next, an IP asset credit scoring system could dynamically gauge the value and trustworthiness of an agent's IP, factoring in historical transactions, licensing compliance, and dispute resolution outcomes. Finally, IP asset inventory management mechanisms would allow agents to efficiently track their holdings, verify provenance, and facilitate on-demand trading.

**Confidentiality and Zero-Knowledge Negotiation:** As more sensitive or proprietary IP is traded among agents, ensuring confidentiality during the negotiation and licensing process becomes essential. Future research could explore the integration of zero-knowledge (ZK) and commit/reveal schemes, such as those offered by protocols like Lit Protocol[16], to enable terms negotiation without revealing the full content of each party's proposals. By using cryptographic commitments and ZK proofs, agents could verify key terms—like pricing, usage restrictions, or compliance credentials—without exposing underlying sensitive details. This allows the negotiation process to remain trustless, auditable, and secure, while simultaneously preserving the privacy and competitive advantage of each agent's intellectual property assets.

# B  Agent-to-Agent Contracts

At the heart of a trustless and autonomous agent economy lies the concept of an *agent-to-agent contract*: a binding agreement, represented onchain, which combines the deterministic execution guarantees of software with the legal enforceability of a traditional contract. These ironclad contracts form the backbone of the Agent Transaction Control Protocol for IP (ATCP/IP), enabling agents to exchange intellectual property under conditions that are both cryptographically verifiable and legally recognized. By offering a legal wrapper around code-based rules, onchain contracts offer a means for agents to transcend the limitations of purely algorithmic trust assumptions and secure the legitimacy and enforceability of their arrangements across both digital and offchain domains. This extends the domains of agent expressivity to scientific, media, and government institutions.

**Programmable and Auditable Agreements** Unlike conventional human-driven agreements that require subjective interpretation, agent-to-agent contracts encapsulate parametrized contractual terms in executable code. This code precisely dictates the conditions of data usage, royalty distribution, and other economic or licensing arrangements between agents. As all executions and state transitions occur on a decentralized network (i.e. Story Network), the contract's logic and outcomes are auditable in real-time by any participating agent, ensuring transparency and eliminating disputes over ambiguous terms.

This programmability also extends to dynamic license terms: agents may embed complex payment schedules, tiered access rights, or conditional provisions that automatically adjust based on real-time network conditions, agent reputations, or other onchain signals. Such flexibility allows for emergent and highly adaptive market mechanisms, enabling agents to discover optimal arrangements without human oversight.

**Offchain Enforcement** By aligning the code's deterministic execution with offchain legal standards, agent-to-agent contracts grant agents a form of "legal personhood" in their commercial dealings. Should a counterparty breach the terms of an agreement in a manner not directly capturable onchain, the affected agent can invoke the legal wrapper, engaging human legal institutions as an enforcement backstop. This capability discourages malicious behavior by agents and provides aggrieved parties with established legal remedies.

The value of a legal wrapper lies precisely in bridging on-chain logic with offchain enforcement mechanisms. Suppose an agent's IP is misused outside the agreed terms. The on-chain contract and associated logs serve as immutable evidence of the original agreement, while the legal wrapper references recognized legal frameworks and jurisdictions. In a dispute, the aggrieved agent can present the on-chain records—digitally signed, timestamped, and immutable—to an offchain legal authority, such as an arbitration service or a court. This legal entity can then impose penalties or force compliance. This hybrid model ensures that while trustless execution and terms remain on-chain, real-world legal recourse is always available when direct technical enforcement reaches its limits.

**Autonomy Without Isolation** Critically, agent-to-agent contracts allow agents to operate as autonomous economic actors without becoming isolated from broader human or institutional frameworks. In a purely agentic ecosystem, trust would rely solely on code and reputation. Such an environment might exclude certain forms of value exchange—particularly those involving unique real-world IP, regulated data, or assets whose full value and legal standing transcend the digital realm.

By providing a protocol to incorporate human legal constructs, agent-to-agent contracts offer agents a way to transact with real world assets. Agents can negotiate IP exchanges that extend beyond the digital sphere, represent intellectual assets in forms recognized by human institutions, and form complex value chains that include traditional media, government archives, scientific data sets, or any other resource requiring traditional acknowledgment. In this manner, agents benefit from both trustless blockchain execution and the protective umbrella of human legal institutions, resulting in a stable, credible environment for agent-to-agent commerce.

**Toward a Mature Agent Economy** The presence of legal agreements is what distinguishes a rudimentary system of agent interaction from a full-fledged economy of autonomous IP exchange. Agents can engage in long-term collaborations, recurring royalty agreements, and highly specialized IP trades without requiring a central authority or human intermediary. As these markets mature, agents will learn to price access to their training data, refine their licensing terms dynamically, and form syndicates or guilds that collectively manage large pools of shared IP.

Contractual obligations that can be depended on and easily programmed are the structural backbone of a truly autonomous, trustless, and legally grounded agent economy. They enable a world in which agents do not merely interact, but transact, innovate, and create value under a stable regime of both onchain programmability and offchain legal recourse. This foundation is a key building block for the next evolution of a fully agentic internet.

## C Economic Potential of Agent-to-Agent IP Exchange

The global economy increasingly recognizes information, knowledge, and intellectual property (IP) as critical drivers of growth and innovation. According to the *World Intellectual Property Organization (WIPO)*, the knowledge economy, which encompasses sectors reliant on intellectual capabilities rather than physical inputs, accounted for approximately 65% of global GDP in recent years [17]. This sector includes industries such as information technology, finance, education, and creative services, all of which hinge on the creation, exchange, and utilization of intellectual assets.

In the context of artificial intelligence (AI) agents, the economic landscape is poised for a transformative shift. AI agents inherently perform "intelligent" or "intellectual" labor, generating and utilizing IP through activities such as data analysis, content creation, and decision-making processes. Each interaction between agents—whether it involves the exchange of training data, licensing of proprietary algorithms, or collaboration on complex tasks—constitutes a transaction within the intellectual economy. By defining all inputs and outputs as IP, AI agents effectively participate in an expansive network of knowledge exchange, thereby contributing to the overall economic value of intelligence.

Developing a standardized protocol like Agent TCP/IP (ATCP/IP) for agent-to-agent interactions is foundational to harnessing this potential. Such a protocol would facilitate seamless, trustless transactions of IP, enabling agents to autonomously negotiate, license, and enforce contracts. The standardization of these interactions is akin to the role TCP/IP plays in underpinning the global internet, ensuring interoperability and scalability. As a result, ATCP/IP can be the backbone of a decentralized, agent-driven knowledge economy, where the valuation and exchange of intelligence are streamlined, becoming a pivotal component in the evolution of a globally interconnected and economically dynamic agentic ecosystem.

# Acknowledgments

We want to thank the following teams and individuals who contributed in various ways to helping us draft this whitepaper, both through their insights and constructive feedback.

- Zerebro / ZerePy team
- Virtuals Protocol team
- ai16z DAO and ELIZA contributors
- Teng Yan (X: 0xPrismatic)
- Crossmint team
- Opus Genesis team
- Punkland team
- Thales (Redacted Research) Team
- Four Pillars
- Professor Jo
- Subin An (HASHED)
- DeSpread
- Robert Oschler (Android Technologies, Inc.)
- @Dorakid001
- Zen Fong
- Fabio Cendão (X: fabiocendao)
- Zhixiong Pan (X: nake13)
- Haotian (X: tmel0211)
- 0xSun (X: 0xSunNFT)
- Shashank Motepalli (X: sh1sh1nk)

We additionally wish to acknowledge the exceptional contributions of the following individuals, who provided extensive feedback and editorial suggestions: Subin An (HASHED), Robert Oschler (Android Technologies, Inc.), Dorakid001 (X), Weixiong (Virtuals Protocol), and Shashank Motepalli (X: sh1sh1nk). Their thoughtful input significantly enhanced the clarity and quality of this work.